\documentclass{article}
\usepackage{amsmath,graphicx}
\usepackage[preprint]{spconf}
\usepackage[dvipsnames]{xcolor}

\usepackage{bm}
\usepackage{multirow}
\usepackage{multicol}
\usepackage{caption}
\usepackage{subcaption}
\usepackage{booktabs}
\usepackage{url}
\usepackage{amssymb}
\usepackage{dsfont}
\usepackage{hyperref}
\usepackage{cite}
\usepackage{comment}
\usepackage{tikz}

\title{SUPERB @ SLT 2022: Challenge on Generalization and Efficiency \\ of Self-Supervised Speech Representation Learning}

\name{%
\begin{tabular}{@{}c@{}}
Tzu-hsun Feng\textsuperscript{1},
Annie Dong\textsuperscript{2},
Ching-Feng Yeh\textsuperscript{2},
Shu-wen Yang\textsuperscript{1},
Tzu-Quan Lin\textsuperscript{1}, \\
\textit{Jiatong Shi}\textsuperscript{3},
\textit{Kai-Wei Chang}\textsuperscript{1},
\textit{Zili Huang}\textsuperscript{4},
\textit{Haibin Wu}\textsuperscript{1},
\textit{Xuankai Chang}\textsuperscript{3}, \\
\textit{Shinji Watanabe}\textsuperscript{3},
\textit{Abdelrahman Mohamed}\textsuperscript{2},
\textit{Shang-Wen Li}\textsuperscript{2}, and \textit{Hung-yi Lee}\textsuperscript{1}
\end{tabular}}
\address{
    \textsuperscript{1}National Taiwan University, Taiwan,
    \textsuperscript{2}Meta, USA,\\
    \textsuperscript{3}Carnegie Mellon University, USA,
    \textsuperscript{4}Johns Hopkins University, USA
}
\newcommand\copyrighttext{%
  \tiny Copyright 2023 IEEE. Published in the 2022 IEEE Spoken Language Technology Workshop (SLT) (SLT 2022), scheduled for 19-22 January 2023 in Doha, Qatar. Personal use of this material is permitted. However, permission to reprint/republish this material for advertising or promotional purposes or for creating new collective works for resale or redistribution to servers or lists, or to reuse any copyrighted component of this work in other works, must be obtained from the IEEE. Contact: Manager, Copyrights and Permissions / IEEE Service Center / 445 Hoes Lane / P.O. Box 1331 / Piscataway, NJ 08855-1331, USA. Telephone: + Intl. 908-562-3966.}
\newcommand\copyrighttextinbottomleft{%
\begin{tikzpicture}[remember picture,overlay]
\node[anchor=south,yshift=20pt] at (current page.south) {\parbox{\dimexpr\textwidth-\fboxsep-\fboxrule\relax}{\copyrighttext}};
\end{tikzpicture}%
}

\copyrightnotice{978-1-6654-7189-3/22/\$31.00~\copyright2023 IEEE}
\begin{document}
\ninept
\maketitle
\copyrighttextinbottomleft
\begin{abstract} 


We present the SUPERB challenge at SLT 2022, which aims at learning self-supervised speech representation for better performance, generalization, and efficiency. The challenge builds upon the SUPERB benchmark and implements metrics to measure the computation requirements of self-supervised learning (SSL) representation and to evaluate its generalizability and performance across the diverse SUPERB tasks. The SUPERB benchmark provides comprehensive coverage of popular speech processing tasks, from speech and speaker recognition to audio generation and semantic understanding. As SSL has gained interest in the speech community and showed promising outcomes, we envision the challenge to uplevel the impact of SSL techniques by motivating more practical designs of techniques beyond task performance. We summarize the results of 14 submitted models in this paper. We also discuss the main findings from those submissions and the future directions of SSL research.
\end{abstract}

\section{Introduction} 
\label{sec:intro}
Today’s commercial speech recognition systems in both academic and industry fields require ever-increasing volumes of text-annotated speech signals for training. The need for massive data in supervised learning hinders the fast advancement of speech processing research. To tackle this issue, self-supervised learning (SSL) \cite{mohamed2022self} has emerged to reduce the dependency on large labeled data sets. SSL utilizes proxy supervised learning tasks (also called pretext tasks) to obtain training data from the tremendous amount of unlabeled corpora available on the web. Researchers explore various pretext tasks to pre-train large neural networks without labels and transfer the pre-trained networks to solve complicated downstream tasks. Recently, SSL has become one of the research mainstreams in speech processing as well as other machine learning communities such as Natural Language Processing (NLP) \cite{kenton2019bert, brown2020language} and Computer Vision (CV) \cite{chen2020simple, grill2020bootstrap}.

Existing SSL research in the speech area centers around learning more powerful pre-trained networks that yield higher accuracy in downstream tasks. SSL techniques have been shown critical to advance research with state-of-the-art (SOTA) results in speech tasks such as automatic speech recognition (ASR) \cite{zhang2020pushing}, automatic speaker verification (ASV) \cite{wang2021fine}, query by examples (QbE), and intent classification (IC) \cite{chen2022unispeech}. Pre-trained networks also show promising performance in low or zero resource scenarios \cite{hsu2021hubert, baevski2020wav2vec}. These studies often evaluate SSL techniques in different benchmarking tasks and datasets, downstream model architectures, and fine-tuning techniques (e.g., fine-tuning entire models or freezing the pre-trained networks). To encourage comparable experiments and a comprehensive understanding of SSL techniques, SUPERB \cite{yang21c_interspeech, tsai2022superb} was introduced to the speech community. SUPERB aims to provide a standard framework to train, evaluate, and compare the efficacy and generalizability of SSL speech networks on 13 speech tasks in recognition, detection, semantics, speaker, paralinguistics, and generation domains.

We observed a growing interest in SUPERB from the speech community. For example, a SUPERB session was organized in The 2nd Workshop on Self-supervised Learning for Audio and Speech Processing @ AAAI 2022\footnote{https://aaai-sas-2022.github.io/} and tutorials about SSL methodologies and their evaluation using SUPERB, were given at ICASSP and NAACL 2022 \cite{lee2022self}. Despite the abundant interest, research still focuses on getting better accuracy, and this focus inevitably leads to using more model parameters, pre-training data, and computation resources. The demand for computation resources is prohibitive for SSL getting wider adoption in academia and production where real-time inference is critical and for more researchers to participate and advance the SSL technology.

The \textit{SUPERB @ SLT 2022: challenge on generalization and efficiency of self-supervised speech representation learning} (denoted as \textbf{the challenge} in the following) is then organized to motivate more and diverse SSL innovation. In the challenge, we establish multiple groups of downstream tasks based on the SUPERB and extended SUPERB-SG \cite{yang21c_interspeech, tsai2022superb} tasks for participants to analyze model capability from different aspects. Diverse metrics beyond accuracy, including memory usage and number of operations, are built to encourage the exploration of efficient SSL techniques with lower memory footprint and computation requirements. Besides, we open two tracks of submission, the public-set track and the hidden-set track, for tuning developed techniques and evaluating the generalizability of techniques, respectively. The goal of the challenge is to 1) continue existing momentum in SSL innovation, and 2) motivate the community to rethink the design criteria and evaluation metrics, and encourage generalizable and efficient SSL techniques beyond task performance. In this paper, we summarize the challenge organization, participating SSL techniques, and the outcomes. 
After the challenge was open, we have gathered 5 different models submitted to the public-set leaderboard and 9 different models submitted to the hidden-set leaderboard. 
Among the models on the hidden-set leaderboard, one has been accepted by Interspeech 2022, 4 submitted papers to SLT 2022, and the other four have not yet been published.
They have enormously different model sizes (from 22.50M to 619.82M) and design targets (achieving SOTA, model compression, more robustness, etc.).
We hope the challenge evangelizes SSL for the speech community.

\section{Background} 
\label{sec:background}
\subsection{Self-supervised learning (SSL) for speech}

Self-supervised learning (SSL) is getting popular in the speech area nowadays, after its wide success in Natural Language Processing (NLP) \cite{kenton2019bert, brown2020language} and Computer Vision (CV) \cite{chen2020simple, grill2020bootstrap}. 
Based on the design of pretext tasks, which are used to obtain proxy supervision tasks for pre-training networks in SSL, SSL techniques in speech can be classified into three categories: generative approaches, contrastive approaches, and predictive approaches. 

For generative approaches, the pretext task is to generate data, i.e., reconstruct the input, based on the masked or corrupted input or to predict future input from the past. Approaches in this category consist of APC \cite{chung2019unsupervised, chung2020generative,chung2020improved, chung20e_interspeech}, DeCoAR \cite{ling2020deep, ling2020decoar}, Mockingjay \cite{liu2020mockingjay,chi2021aalbert}, TERA \cite{liu2021tera}, MPC~\cite{jiang2019improving,jiang2021further,yue2021pMPC} , speech-XLNet~\cite{song20d_interspeech}, NPC~\cite{liu21l_interspeech}, and PASE+ \cite{pascual2019learning,ravanelli2020multi}. Pretext tasks for contrastive approaches are designed to learn a latent space representation, where distance is minimized between the anchor and positive examples while maximized between the anchor and negative. Positive and negative examples are often created by utilizing the distance between input frames and data augmentation. Representative works in contrastive approaches include CPC \cite{oord2018representation}, wav2vec~\cite{schneider2019wav2vec, Baevski2020vq-wav2vec}, and wav2vec 2.0~\cite{baevski2020wav2vec}. 
Inspired by BERT in NLP, predictive models, such as HuBERT and WavLM \cite{hsu2021hubert, chen2022wavlm}, aim good representations from unmasked input and infer discrete tokens at masked positions. 
Given the growing popularity of SSL in the speech area, an overview paper \cite{mohamed2022self} and a tutorial \cite{lee2022self} have been put to review recent works in this direction and attract more researchers to contribute.

\subsection{Benchmarking SSL techniques in speech}
The efficacy of SSL techniques is usually evaluated in the following two-step setting. First, networks are pre-trained with pretext tasks designed in SSL techniques. Then the pre-trained networks are appended with optional prediction layers and fine-tuned with various downstream tasks. The fine-tuned results are evaluated towards downstream tasks to measure the efficacy of SSL techniques and the quality of their learned networks. There are established benchmarks for evaluating SSL in common settings, such as GLUE \cite{wang2018glue} in NLP, VTAB \cite{zhai2019large} in CV, and CH-MARL in Multimodal \cite{sharma2022ch}. However, researchers in the speech area often adopted more diverse settings \cite{mohamed2022self}. The design of prediction layers and fine-tuning methods are different across works and downstream tasks ranging from ASR and ASV to speech translation (ST) and Speech enhancement (SE).

Due to the diversity in evaluation settings, there are increasing efforts to establish common benchmarking settings with shared downstream tasks and datasets. SUPERB \cite{yang21c_interspeech, tsai2022superb} consolidates settings of prediction layer architectures, hyperparameter spaces, and tunable parameters to provide the SSL community with a benchmarking platform. SUPERB allows researchers to plug in their pre-trained networks and evaluate on 13 different speech tasks by leveraging s3prl toolkit\footnote{https://github.com/s3prl/s3prl, this is a toolkit independent from SUPERB benchmark and challenge, but is fully capable to evaluate them.}. LeBenchmark \cite{evain21_interspeech} provides a benchmark to understand the performance of SSL techniques in French. ZeroSpeech \cite{dunbar2020zero} is a challenge to build speech and language understanding with zero expert resources and is also used for analyzing the quality of SSL models \cite{tjandra20_interspeech, niekerk20b_interspeech}. These efforts help researchers understand SSL techniques in downstream performance. Since SUPERB has diverse speech processing tasks and uses both popular publicly available datasets and  newly generated datasets in those tasks. In this challenge, we extend SUPERB and encourage the community to think further for efficient use of computation resources while building powerful techniques.

\begin{figure*}
    \centering
    \includegraphics[width=1.\linewidth]{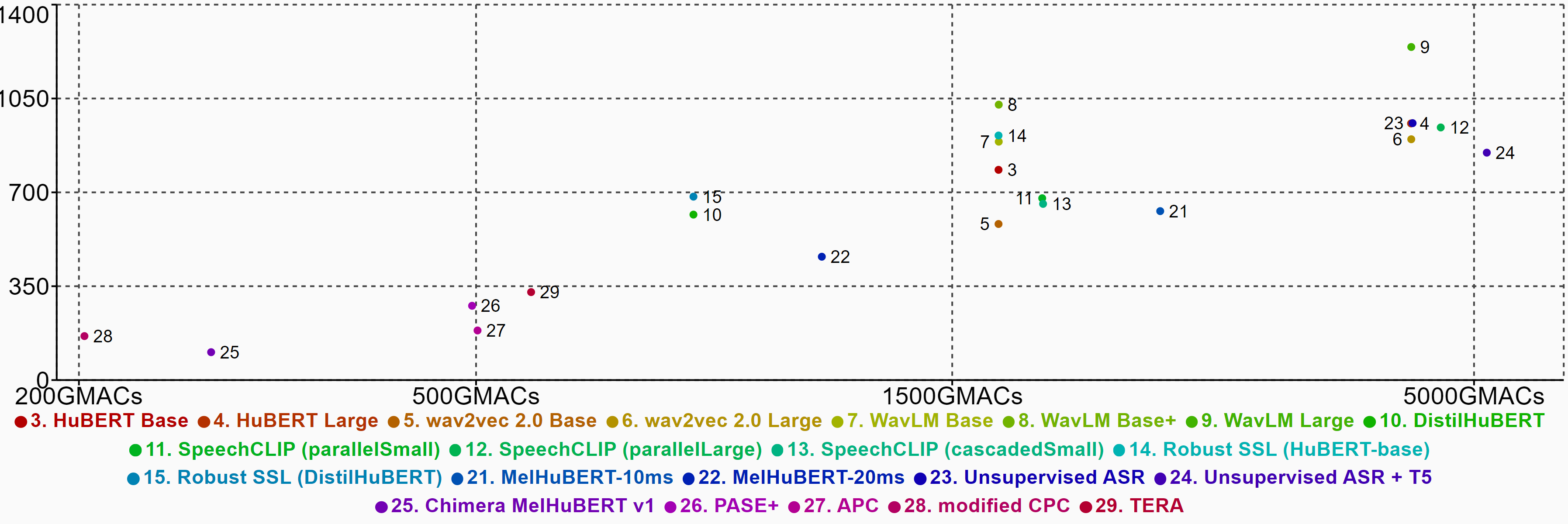}
    \caption{This figure shows the MACs (X-axis, in log scale) versus SUPERB score $\mathrm{superb_s}$ (Y-axis) on the hidden-set track, including all except FBANK models below, which have run ten tasks.}%
    \label{fig:result}
\end{figure*}

\section{Challenge overview} 
\label{sec:overview}

This challenge benchmarks the efficacy of self-supervised learning (SSL) speech networks in various types of downstream tasks and its computation requirements. To keep the cost affordable and accelerate the development iteration, the challenge chooses Phoneme Recognition (PR), Speaker Identification (SID), Emotion Recognition (ER), Automatic Speech Recognition (ASR), Query-by-Example (QbE), Automatic Speaker Verification (ASV), Speaker Diarization (SD), Source Separation (SS), Speech Enhancement (SE), and Speech Translation (ST) from the SUPERB and extended SUPERB-SG \cite{yang21c_interspeech, tsai2022superb} tasks. Researchers participate in the challenge by submitting their SSL pre-trained (i.e., upstream) model. Following the same evaluation framework introduced in them, we, the challenge organizers, extract \textbf{multiple frozen hidden states} from the pre-trained SSL model by default and then train a learnable \textbf{weighted-sum} over the hidden states along with the downstream model with predefined architecture task-by-task. Participants are allowed to apply some task-specified pre/post-processing on their upstream model for better adaptation to each dwnstream task.

\subsection{Metrics}
\label{sec:metrics}
We use two types of metrics to measure the generalizability and computation requirements of SSL speech networks. For the former one, we proposed SUPERB score ($\mathrm{superb_s}$) as an overall metric to show how well each SSL technique performs in all downstream tasks. For the latter one, we choose the theoretical multiply-accumulate operation (MACs) and the number of parameters (Params) as indicators for time and space cost.

\subsubsection{Generalizability Metrics}
\label{ssec:pmetrics}
Let $s_{t,i}$ be the $i$th metrics for task $t$, $s_{t,i}(u)$ be the corresponding score of upstream model $u$, $T$ be the set of tasks, and $I_t$ be the set of metrics for task $t$. We first convert all scores to the same scale by linear interpolation with respect to baseline and SOTA. If a task has more than one metric, we also apply an intra-task average. Then the last step is doing an inter-task average and multiplying 1000.

$$\mathrm{superb_s}(u) = \frac{1000}{|T|}\Sigma_t^T \frac{1}{|I_t|}\Sigma_i^{I_t} \frac{s_{t,i}(u) - s_{t,i}(\mathrm{baseline})}{s_{t,i}(\mathrm{SOTA}) - s_{t,i}(\mathrm{baseline})}$$
To keep the value of SUPERB score static, we obtain the baseline and SOTA from the SUPERB leaderboard snapshot on October 15, 2021, which is the launch of leaderboard. $\mathrm{superb_s}$ is designed to provide a comprehensive view of model capability and take the difficulty of each task into consideration. If the performances of the SOTA and baseline are close on a task, then a little improvement should be more valuable than a task has a large difference.

\subsubsection{Efficiency Metrics}
\label{ssec:emetrics}
Taking advantage of DeepSpeed\footnote{https://github.com/microsoft/DeepSpeed}, an existing deep learning optimization library including a function for profiling neural networks, we use it as the backbone and customize it for profiling networks in this challenge. We implement the estimation of MACs at the software level by wrapping each function / operator in pytorch with an approximation formula, regardless of the real implementations at the hardware level. For feasibility, we select 32 real audios from LibriSpeech \cite{panayotov2015librispeech} test-clean split, chosen equally from short to long. More implementation detail can be found at our fork repository\footnote{https://github.com/B06901052/DeepSpeed/tree/superb-challenge}.

\subsection{Dataset}
\label{sec:dataset}
To facilitate the development of SSL techniques and fair comparison over challenge submissions, we choose to have two datasets for each task. One is public-available, and representative in the corresponding speech processing task, e.g., LibriSpeech \cite{panayotov2015librispeech} used in ASR and PR. The other is newly created by us and held out as hidden-set. The differences between the two datasets are controlled to be only in 1) recording conditions, 2) spoken content / text scripts, 3) speakers, and 4) the amount of labeled data. To build the hidden datasets, we collected English text from selected existing corpora and split text into sentences. We worked with LxT\footnote{https://www.lxt.ai/} and recruited 60, gender balanced human speakers to read sentences and record the audio. All collected data is processed into subsets of audio between 1 to 5 hours for train, dev, test split of each downstream task. LxT also translated provided text to German for our en-to-de Speech Translation task.

\subsection{Challenge participation}
\label{sec:participation}

We provide two submission tracks for participants to choose from. The public-set track is mainly for development, demonstrating the task designs of our challenge. Participants can tune the hyperparameters and apply any fine-tuning techniques to reach the best performance of their models. Submissions to public track are made by uploading the prediction files. Participants cannot access data in the hidden-set track, and they have to submit their pre-trained model checkpoints and model architecture. We evaluate submissions with a fixed training procedure and sweep over several default learning rates defined by ourselves, which will be a subset of $\{1,0.1,...,10^{-6},10^{-7}\}$ for each task. To lower the bar of participation and encourage more diverse aspects of efforts, we also allow only evaluation on a subset of tasks, and participants can control whether to present their results on the online leaderboard or not.

As an online challenge, we are capable of demonstrating results in more comprehensive ways. The classic table show the score of each task. The scatter chart (Fig.~\ref{fig:result}) compares efficiency metrics with performance metrics, and can select which one to use by users, visualizing the trade-off between them. The radar chart (Fig.~\ref{fig:compare}) enables fast performance comparison of models in each task.

\begin{table*}[ht]
\centering
\caption{Evaluating various SSL representations on various downstream tasks in the hidden-set track. The toplines are the SOTA methods from the SUPERB leaderboard snapshot on October 15, 2021, which is the leaderboard launch date. The numbers are collected with public-available checkpoints or codes, and we welcome researchers to re-submit the results to our online leaderboard.
In this table, we also put FBANK, HuBERT-base, HuBERT-large, wav2vec 2.0-base, and wav2vec 2.0-large for comparison.}
\resizebox{1.0\textwidth}{!}{
\begin{tabular}{|l||r|r|r|r|r|r|r|r||r|r|r|r||r|r|r|} 
\hline
\cline{2-16}
 & \multicolumn{1}{c|}{PR} & \multicolumn{1}{c|}{SID} & \multicolumn{1}{c|}{ER} & \multicolumn{1}{c|}{ASR (w/o LM)} & \multicolumn{2}{c|}{QBE} & \multicolumn{1}{c|}{ASV} & \multicolumn{1}{c||}{SD} & \multicolumn{1}{c|}{SS} & \multicolumn{2}{c|}{SE} & \multicolumn{1}{c||}{ST} & \multicolumn{1}{c|}{Generalizability} & \multicolumn{2}{c|}{Computation Requirements} \\ 
\cline{2-16}
 & PER$\downarrow$ & Acc$\uparrow$ & Acc$\uparrow$ & WER$\downarrow$ & MAP$\uparrow$ & EER$\downarrow$ & EER$\downarrow$ & DER$\downarrow$ & SI-SDRi$\uparrow$ & STOI$\uparrow$ & PESQ$\uparrow$ & BLEU$\uparrow$ & $\mathrm{superb_s}$ & MACs (G) & Params (M) \\ 
\hline \hline
1.baseline (FBANK) & 81.66 & 48.17 & 46.98 & 91.54 & 12.72 & 35.98 & 24.04 & 13.40 & 2.85 & 84.46 & 1.5300 & 2.32 & 0 & 0.479 & 0 \\ 
\hline
2.topline (previous SOTA, marked with *) & 18.22 & 80.25 & 60.99 & 27.06 & 49.06 & 16.55 & 9.81 & 9.10 & 7.30 & 85.29 & 1.5694 & 20.01 & 1000 & - & - \\
\hline
3.HuBERT-base & 19.19 & 70.33 & 60.16 & 37.25 & *49.06 & *16.55 & 13.92 & 9.45 & 5.98 & 84.77 & 1.5392 & 15.53 & 784 & 1669 & 94.70 \\ 
\hline
4.HuBERT-large & *18.22 & 80.00 & 64.84 & *27.06 & 31.00 & 33.05 & *9.81 & *9.10 & *7.30 & *85.29 & 1.5676 & *20.01 & 957 & 4324 & 316.61 \\ 
\hline
5.wav2vec 2.0-base & 24.50 & 75.58 & 52.20 & 48.85 & 32.08 & 28.51 & 13.48 & 11.09 & 5.73 & 84.47 & 1.5250 & 14.81 & 582 & 1669 & 95.04 \\
\hline
6.wav2vec 2.0-large & 22.55 & *80.25 & *60.99 & 29.93 & 39.20 & 22.48 & 10.38 & 10.22 & 6.87 & 85.06 & *1.5694 & 18.50 & 898 & 4326 & 317.39 \\
\hline
7.WavLM-base & 19.01 & 71.17 & 54.67 & 36.46 & \textbf{56.49} & \textbf{12.59} & 13.51 & 10.29 & 8.67 & 85.27 & 1.5710 & 16.46 & 889 & 1670 & 94.38 \\ 
\hline
8.WavLM-base+ & \textbf{15.29} & 83.08 & 57.42 & 31.47 & 56.31 & 15.05 & 12.14 & 9.85 & \textbf{9.15} & 85.61 & 1.5763 & 19.34 & 1027 & 1670 & 94.38 \\ 
\hline
9.WavLM-large & 16.80 & \textbf{92.75} & 66.21 & \textbf{24.48} & 50.97 & 16.93 & \textbf{7.97} & \textbf{8.73} & 9.13 & \textbf{85.96} & 1.5999 & \textbf{22.93} & \textbf{1242} & 4326 & 315.45 \\ 
\hline
10.DistilHuBERT & 35.83 & 74.75 & 56.59 & 64.09 & 44.08 & 19.08 & 12.33 & 10.62 & 5.45 & 84.61 & 1.5323 & 10.55 & 617 & 826 & 27.03 \\ 
\hline
11.SpeechCLIP (parallel small) & 19.98 & 64.00 & 60.99 & 37.65 & 49.13 & 16.74 & 14.30 & 11.22 & 5.52 & 84.48 & 1.5276 & 14.96 & 678 & 1846 & 109.27 \\ 
\hline
12.SpeechCLIP (parallel large) & 15.76 & 85.17 & 61.81 & 27.46 & 31.00 & 33.05 & 9.88 & 9.37 & 7.43 & 85.36 & 1.5549 & 20.64 & 942 & 4630 & 342.59 \\ 
\hline
13.SpeechCLIP (cascaded small) & 19.80 & 57.33 & 57.97 & 37.57 & 49.13 & 16.74 & 14.06 & 10.68 & 5.53 & 84.43 & 1.5341 & 15.02 & 657 & 1850 & 109.49 \\ 
\hline
14.Robust SSL (HuBERT-base) & 18.61 & 81.25 & 60.16 & 36.10 & 55.19 & 14.10 & 12.30 & 10.49 & 6.46 & 85.57 & 1.5644 & 15.22 & 912 & 1669 & 94.70 \\ 
\hline
15.Robust SSL (DistilHuBERT) & 30.87 & 79.42 & 56.04 & 61.13 & 45.92 & 19.58 & 11.44 & 10.89 & 4.92 & 85.08 & 1.5556 & 9.95 & 684 & 826 & 27.03 \\ 
\hline
16.adding silence (HuBERT-base, front 1/10) & - & - & - & - & - & - & 12.98 & - & - & - & - & - & - & 1848 & 94.70 \\ 
\hline
17.adding silence (HuBERT-large, front 1/10) & - & - & - & - & - & - & 9.61 & - & - & - & - & - & - & 4787 & 316.61 \\ 
\hline
18.Sequence reduction (w2v2u, last layer) & 37.84 & 72.25 & 58.79 & - & 38.02 & 20.27 & 14.43 & - & - & - & - & - & - & 617 & 24.80 \\ 
\hline
19.Sequence reduction (w2v2u, all layers) & 38.73 & 72.92 & 52.47 & - & 46.18 & 17.01 & 13.51 & - & - & - & - & 7.26 & - & 626 & 28.35 \\ 
\hline
20.Sequence reduction (l25, all layers)& 38.09 & 72.42 & 57.42 & - & 43.89 & 16.59 & 13.45 & - & - & - & - & 7.18 & - & 633 & 28.35 \\
\hline
21.MelHuBERT-10ms & 31.43 & 63.92 & 48.90 & 50.04 & 36.19 & 27.72 & 17.00 & 11.04 & 5.83 & 85.07 & \textbf{1.6236} & 9.99 & 630 & 2424 & 90.20 \\ 
\hline
22.MelHuBERT-20ms & 25.98 & 53.25 & 49.45 & 48.18 & 35.68 & 29.26 & 16.89 & 11.27 & 4.85 & 84.61 & 1.5433 & 11.55 & 460 & 1110 & 90.20 \\
\hline
23.Unsupervised ASR & 17.22 & 91.17 & 65.11 & 31.11 & 40.51 & 21.99 & 9.60 & 10.26 & 7.20 & 85.07 & 1.5500 & 18.33 & 958 & 4339 & 320.18 \\ 
\hline
24.Unsupervised ASR + T5 & 18.14 & 80.92 & \textbf{66.48} & 36.25 & 40.51 & 21.99 & 11.39 & 11.11 & 6.70 & 84.79 & 1.5402 & 17.39 & 848 & 5149 & 619.82 \\ 
\hline
25.Chimera MelHuBERT v1 & 52.17 & 50.50 & 55.22 & 86.45 & 27.36 & 27.13 & 16.00 & 12.47 & 2.12 & 82.92 & 1.4921 & 5.69 & 104 & 271 & 22.50 \\
\hline
\end{tabular}
}
\label{table:exp}
\end{table*}

\section{Submissions} 
\label{sec:results}

\subsection{Submission to Public Leaderboard}

This subsection briefly introduces the models submitted to the public-set track.
The models in this subsection are not submitted to the hidden-set track, so their results are not in Table~\ref{table:exp} and Fig.~\ref{fig:result}, except WavLM series and DistilHuBERT, which are evaluated on the hidden-set by the organizers for comparison. 

\textbf{WavLM series (WavLM Base, WavLM Base+, WavLM Large)}~\cite{chen2022wavlm}: 
WavLM used the same pretext task as HuBERT~\cite{hsu2021hubert}.
The WavLM framework proposed an utterance mixing strategy where partially overlapped signals from different speakers are constructed to augment the training data. 
The pre-training data used for WavLM extends the 60k hours of pre-trained audio used for HuBERT and wav2vec 2.0 to reach a total of 94k hours of audio. 

\textbf{data2vec Large}~\cite{data2vec}\footnote{The model is publicly available. It is not submitted by the original authors.}: 
data2vec predicts contextualized latent input representations given the masked view of the input. 
The data2vec approach was shown to work well for speech representation learning and visual and text representations~\cite{data2vec}. 


\textbf{FaST-VGS+}~\cite{FaSTVGSplus}:
FaST-VGS+ is an SSL model that learns to associate raw speech waveforms with semantically related images.
It is learned in a multi-task fashion with a masked language modeling objective in addition to the visual grounding objective.
It uses wav2vec 2.0 base as its initialization. 

\textbf{DistilHuBERT}~\cite{chang2021distilhubert}:
 DistilHuBERT is trained with a teacher-student learning framework with knowledge distillation. 
 The student network consists of a subnet followed by some prediction heads, where the subnet is constructed by reducing the number of transformer encoder layers of the HuBERT teacher model. 
 Given an input speech utterance, prediction heads predict the hidden layer representations of some specific layers of the teacher model. 

\textbf{LightHuBERT series (LightHuBERT Stage 1 and LightHuBERT Small)}~\cite{LightHuBERT}: 
LightHuBERT is a model compression framework that consists of a once-for-all Transformer, a contextualized latent representation distillation objective, and a two-stage training strategy. 
Stage one trains the largest architecture of the once-for-all Transformer from scratch via the loss function of the pre-training distillation. 
Stage two implements the once-for-all training on the supernet initialized by distilled weights. 

\begin{figure}
    \centering
    \includegraphics[width=1.0\linewidth]{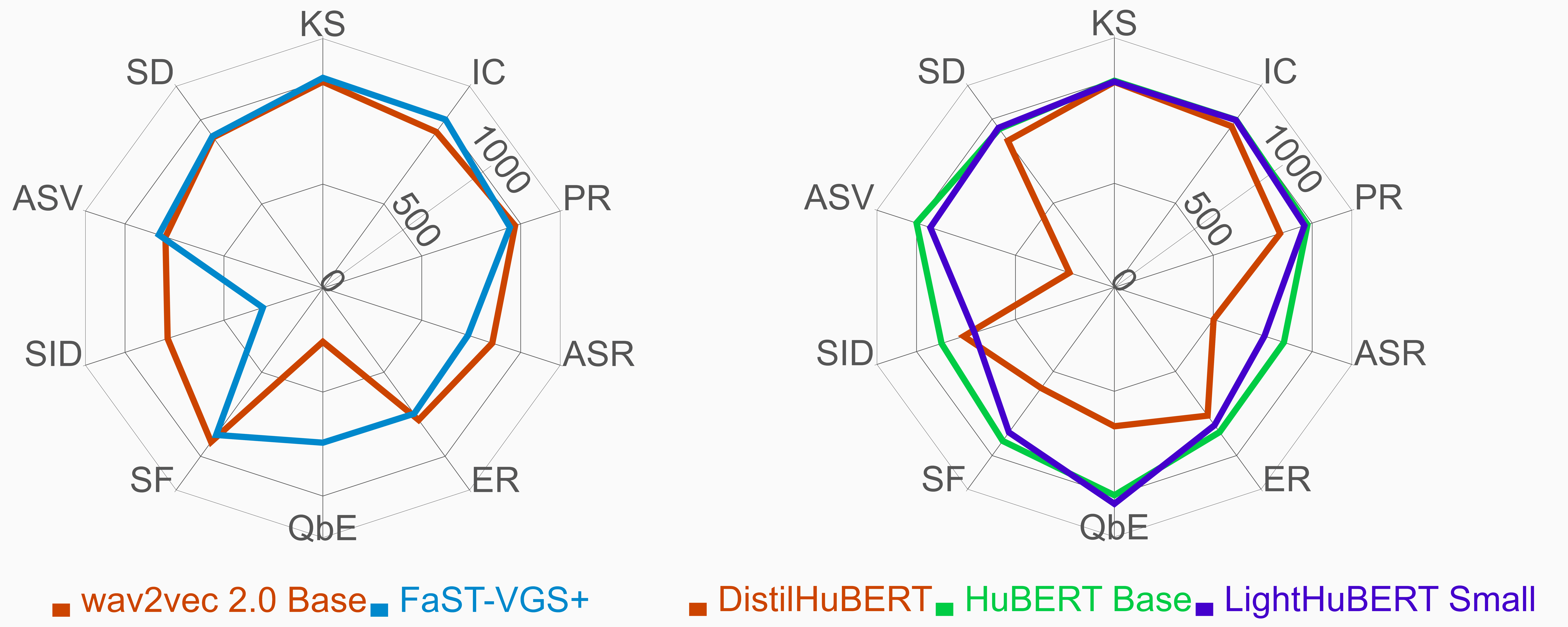}
    \caption{Model performance comparison by radar chart on the public-set. The value is component of SUPERB score from each task before doing inter-task average.
    Left: Comparison of FaST-VGS+ and its initialization, wav2vec 2.0. 
    Right: Comparison of HuBERT and its compressed version, DistilHuBERT and LightHuBERT small.
    We compare DistilHuBERT and LightHuBERT because they have close model sizes.}
    \label{fig:compare}
\end{figure}

\begin{figure}
    \centering
    \includegraphics[width=1.0\linewidth]{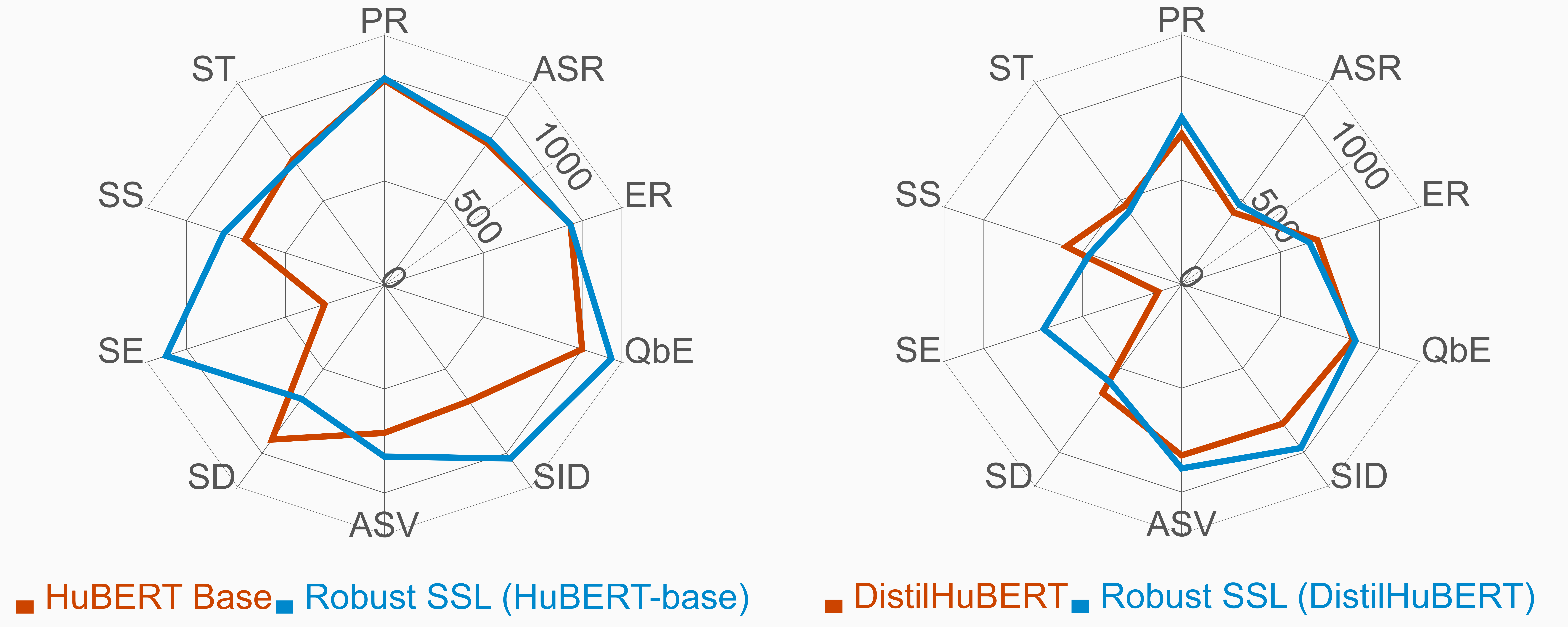}
    \caption{Model performance comparison by radar chart on the hidden-set. The value is component of SUPERB score from each task before doing inter-task average.
    Left: Comparison of HuBERT-base and Robust SSL (HuBERT-base).
    Right: Comparison of DistilHuBERT and Robust SSL (HDistiluBERT).
    }
    \label{fig:compare_robust}
\end{figure}

\subsection{Submission to Hidden-set Track}

Here we introduce the SSL models submitted to the hidden-set track.
The models without reference are not yet officially published or submitted to any conference when writing this paper.

\textbf{SpeechCLIP}~\cite{SpeechCLIP}: 
SpeechCLIP is a framework bridging speech and text through images to enhance speech SSL models by leveraging Contrastive Language-Image Pre-training~(CLIP), a model pre-trained to align parallel image-text data~\cite{radford2021clip}.
SpeechCLIP has two versions: \textit{parallel} and \textit{cascaded}.
The parallel SpeechCLIP aligns speech and CLIP image encoders and implicitly bridges speech and text representations since CLIP's image and text encoders are well-align.
The cascaded SpeechCLIP cascades CLIP's text encoder on top of the speech encoder, forcing the model to output subword embeddings.
All the SpeechCLIP models add additional layers on top of HuBERT, and only the additional layers are learned with CLIP.
For the SpeechCLIP submissions, the model with "small" in its name is based on HuBERT-base, while "large" means based on HuBERT-large.
 

\textbf{Robust SSL}~\cite{RobustHuBERT,RobustDistilHuBERT}:
Pre-training SSL models with target domain data is an intuitive way to adapt SSL models to another domain \cite{hsu2021robust}. 
This submission continuously pre-trained the HuBERT-base model with distorted speech data for some additional steps to enhance robustness. 
Distorted speech is generated by applying both additive distortions and non-additive distortions, and the details are in~\cite{RobustDistilHuBERT}.

\textbf{Robust SSL with Knowledge distillation}~\cite{RobustDistilHuBERT}: 
This SSL model uses the teacher-student framework as DistilHuBERT~\cite{chang2021distilhubert}.
To overcome the problem that DistilHuBERT is especially vulnerable to distorted speech, the teacher model performs continual training as in the last paragraph so that the student model would have a more robust target to learn with, and during knowledge distillation, the teacher and student model have different distorted inputs.

\textbf{ccc-wav2vec 2.0}~\cite{cccw2v2}\footnote{We thanks the authors for providing the model description. 
The authors plan to submit the model but cannot do it before the submission of this paper due to time limitations, so the results of the model are not in Table~\ref{table:exp} and Fig.~\ref{fig:result}.}: 
 ccc-wav2vec 2.0 uses clustering and an augmentation-based cross-contrastive loss as its self-supervised objective. 
 The Cross-Contrastive loss is computed between the encoder output of the original sample and the quantizer output of its augmentation, and vice-versa, bringing robustness to the pre-training strategy. 
 ccc-wav2vec 2.0 achieves up to 15.6\% and 12.7\% relative WER improvement over the baseline wav2vec 2.0 on the test-clean and test-other sets, respectively, of LibriSpeech, without the use of any language model. 
It also achieves up to 14.9\% relative WER improvement over the baseline wav2vec 2.0 when fine-tuned on Switchboard data.


\textbf{Adding silence}~\cite{AddSil}: 
It has also been shown that in the SID task (public-set track) of SUPERB, if the silence ratio is less than  5\%, performance will be reduced by about 30\% to 50\% compared to other cases~\cite{AddSil}. 
The observation inspired a straightforward way to improve the accuracy of speaker-related tasks -- adding silence to the utterances without sufficient silence.
This submission modifies the preprocessing process by padding silence in front of the utterances before HuBERT-base and HuBERT-large models extract the representations.
The lengths of silence are 1/10 of the whole utterances. 
This submission only modifies the preprocessing without changing the existing SSL models. 
The submissions are only evaluated on ASV because this approach is only designed to improve speaker-related tasks.


\textbf{MelHuBERT}: 
MelHuBERT is a replication study of HuBERT on Melspectrogram with simplified HuBERT loss. The only difference in model architecture is that MelHuBERT does not have convolutional feature extractor at the beginning. Instead, it directly takes Melspectrogram as input. There are two variants with different frame period, MelHuBERT-10ms and MelHuBERT-20ms. Despite that MelHuBERT-10ms has more MACs comparing to HuBERT due to input sequence length, MelHuBERT-20ms successfully reduce the MACs by 33\% by removing the convolutional feature extractor. To further reduce the computational overhead during pre-training, MelHuBERT-10ms and MelHuBERT-20ms only use a smaller batch size of 32.

\textbf{Sequence reduction}~\cite{SeqCompress}: 
The paper~\cite{SeqCompress} investigates variable-length subsampling to reduce the sequence length along the time axis to reduce the computational cost.
In variable-length subsampling, a sequence of boundaries is first detected by Continuous Integrate-and-Fire (CIF)~\cite{CIF}; vectors within each segment are pooled and passed to the subsequent layers.
The paper also investigates using different approaches to guide the learning of CIF.
``w2v2u'' and ``l25'' in the submissions refer to different guidance approaches.
Please refer to~\cite{SeqCompress} for details. 
``last layer'' means only the last layer output is used in downstream tasks, while ``all layers'' means the representations from all layers are weighted sum. 
As the sequence reduction approaches change the sequence lengths, they cannot be evaluated on some tasks.




\textbf{Chimera MelHuBERT v1}: 
This model combines Robust SSL (DistilHuBERT) and MelHuBERT. 
MelHuBERT has demonstrated the computation requirements of taking Melspectrogram as input while reducing the CNN feature extractor. 
Chimera MelHuBERT v1 further incorporates this technique in Robust SSL. 
Specifically, in the robust distillation framework, the teacher model is the robust HuBERT, and the student model is the DistilHuBERT with Melspectrogram as input. 
Therefore, during distillation, only two transformer layers are trained on top of the Melspectrogram with the robust distill loss as the objective. As a result, Chimera MelHuBERT v1 can achieve low MACs and few trainable parameters.

\textbf{Unsupervised ASR (wav2vec-u 2.0)}:
Hidden representations from self-supervised models have been used in the unsupervised ASR task \cite{lin2022analyzing, liu2022towards, baevski2021unsupervised}. 
The submission considers an adversarial-trained unsupervised ASR model (i.e., wav2vec-u 2.0) as an SSL model.
The phoneme posteriorgram is an additional feature for downstream tasks. 
As phoneme posteriorgram and wav2vec2's representations do not have the same time resolution, we upsample the posteriorgram, which is done just by repeating the vectors, by a factor of three to match the 20ms frame-shift for wav2vec2. 
The predicted phoneme posteriorgram is concatenated with the wav2vec2 feature as the downstream models' input. 
The unsupervised ASR model is trained on Librispeech-960 hours with unpaired text from the Librispeech language modeling corpus. 
Note that the unsupervised ASR model is trained without the auxiliary K-means loss as its original version to improve stability \cite{liu2022towards}. 

\textbf{Unsupervised ASR + Phoneme T5}:
Unsupervised ASR transcribes speech signals into phoneme sequences. 
This submission leverages the encoder of phoneme-based T5\footnote{\url{https://huggingface.co/voidful/phoneme_byt5}}, a variant of T5 that takes phonemicized text as input.
Phoneme-based T5 encoder takes the output of unsupervised ASR as input and generates a sequence of representations.
The representations generated from the phoneme-based T5 encoder are concatenated with the wav2vec2 feature and phone posteriorgram as an additional feature. 
The model's target is to generate semantic representation from the phoneme-based T5 encoder to improve the spoken language understanding tasks. 
Only the last hidden representation from the phoneme-based T5 encoder is applied for the additional feature.

In the following, we summarize the work that modified the upstream models used in the SUPERB benchmark. 
The results in these papers cannot be compared with the submission in Table~\ref{table:exp} because they have different downstream models.

\textbf{Adapter}~\cite{speechadapter}: 
This study aims to explore efficient tuning methods for speech self-supervised learning. 
Adapters are lightweight modules inserted into SSL models.
In downstream tasks, the parameters of SSL models are frozen, and only the adapters are trained. 
The study shows that with an adapter, the performance parity can be achieved with over 90\% parameter reduction.
The study further finds that the Houlsby adapter~\cite{houlsby2019parameter} is the most efficient in the trade-off between performance and the number of parameters, and the Weighted sum strategy used in the SUPERB challenge is a very suitable efficient method to use in SSL speech tasks.

\textbf{Correlation Pooling}~\cite{ChannelSpeaker}: 
For utterance-level classification tasks in SUPERB (e.g., SID, ER, etc.), the default downstream model in the challenge is to aggregate the speech representations across time by mean pooling.
Correlation pooling~\cite{ChannelSpeaker}, which extracts correlations between the coefficients of the representations, shows improvements over mean pooling. 


\subsection{Discussion of Hidden-set Track Results} 

Here we summarize the general observations in the challenge, including both the public-set and hidden-set tracks. 
The results of hidden-set track are shown in Table~\ref{table:exp} and Fig.~\ref{fig:result}.

\textbf{The designs of the submitted models vary a lot}. 
WavLM series uses more pre-training data and new augmentation methods.
 data2vec is a new self-supervised objective.
  FaST-VGS+ and SpeechCLIP improve SSL with visual information associated with audio.
  DistilHuBERT, LightHuBERT, MelHuBERT compress the SSL models to reduce the number of parameters, while Sequence reduction shortens the sequence to reduce the computation. 
    Robust SSL (with knowledge distillation) and ccc-wav2vec 2.0 improve the robustness of SSL models.
  ``Adding silence'' modifies the audio preprocessing pipeline. 
Unsupervised ASR (+ Phoneme T5) leverages text unpaired with audio.
Chimera MelHuBERT is the integration of different approaches. 

\textbf{The submissions have a vast range of MACs and network parameters}. Among all the submissions, Chimera MelHuBERT v1 is the model with the least MACs (271G) and network parameters (22.50M). Unsupervised ASR + T5 has the most MACs (5149G) and network parameters (619.82M).

\textbf{On the SUPERB public leaderboard, WavLM Large achieved the SOTA on all the tasks}, except ASR (ranked 2nd place), while data2vec achieved the SOTA on ASR.
On the hidden-set, considering all the SSL models in terms of $\mathrm{superb_s}$ and their computation requirements in Fig.~\ref{fig:result}, for the group with larger MACs, WavLM is still the best; for the small models, ``Robust SSL (DistilHuBERT)'' is the best.
Comparing SSL models with FBANK, all the submissions outperformed FBANK, except in some cases in SS and SE.
The observation shows that SSL models are beneficial for most speech processing tasks.
However, we need more study on leveraging the SSL models effectively in tasks involving speech generation like SS and SE.

\textbf{The results of incorporating visual information are mixed}. 
The comparison of FaST-VGS+ and its initialization wav2vec 2.0 on the public-set track is in Fig.~\ref{fig:compare}.
FaST-VGS+ is especially strong on KS and IC.
On KS and IC, it outperforms other models, except WavLM Base+ and WavLM Large, which use much more pre-training data than FaST-VGS+ (1.7k hours v.s. 94k hours). 
But FaST-VGS+ especially degrades the performance of SID.
On the hidden-set track, for the SpeechCLIP models, when the models have the same size, the parallel model and the cascade model have comparable $\mathrm{superb_s}$ (``cascade small'' v.s. ``parallel small''). 
The SpeechCLIP models labeled with ``small'' adds extra layers on top of HuBERT-base, but their performance is inferior to HuBERT-base for almost all tasks, especially the speaker-related tasks, SID, ASV, and SD, with the same trend as FaST-VGS+.
Increasing the model size of SpeechCLIP improves its performance remarkably (``parallel small'' v.s. ``parallel large'').
We found that SpeechCLIP (parallel large) outperformed HuBERT-large on SID, PR, and ASR, achieved a very low PR error rate, only worse than WavLM-base+.
We see the benefit of incorporating visual information during pre-training on some downstream tasks, but how to leverage visual information to improve all downstream tasks requires further study\footnote{Since FaST-VGS+ does not submit to the hidden-set track, and SpeechCLIP does not submit to the public-set, we cannot compare SpeechCLIP and FaST-VGS+ side-by-side.}.

\textbf{On the public-set track, DistilHuBERT reduces HuBERT-based’s size by 75\% while retaining the performance on some downstream tasks}, but compared with HuBERT-based, the performance of DistilHuBERT remarkably degrades on ASR and ASV. LightHuBERT small achieves comparable performance to the teacher model (HuBERT-base) in most tasks with a parameter reduction of 29\%. 
The comparison of HuBERT, DistilHuBERT, and LightHuBERT are in Fig.~\ref{fig:compare}. 
On the hidden-set track, the three submissions regarding sequence reduction approaches have different performances on ER and QbE, but it is hard to conclude which one is the best.
The sequence reduction submissions have approximately the same parameter numbers as DistilHuBERT while decreasing the MACs, but their performances on most of the tasks also decrease compared with DistilHuBERT.
The results show that we still need more study to reduce representation sequence effectively. 
Sequence reduction is a promising new research direction to reduce the computational cost of the speech SSL models because its reduction is orthogonal to the existing compression models like  DistilHuBERT and LightHuBERT and can integrate with them. 

\textbf{The comparison of the robust SSL models and their counterparts on the hidden-set track} are shown in Fig.~\ref{fig:compare_robust}. 
In summary, robust SSL models improve the $\mathrm{superb_s}$ in the challenge (Robust SSL (HuBERT-base) v.s. HuBERT-base, Robust SSL (DistilHuBERT) v.s. DistilHuBERT), and they are especially good at SE and SID.
Although the testing set in the challenge does not have speech distortion like the original papers proposing the robustness approaches~\cite{RobustHuBERT,RobustDistilHuBERT}, making the DistilHuBERT and HuBERT-based more robust by training with distorted speech still increases their $\mathrm{superb_s}$ in the challenge.

\textbf{On the hidden-set track, adding silence to the front of HuBERT improves the performance of ASV}. The results support the hypothesis that HuBERT uses the representations corresponding to silence to store speaker information~\cite{AddSil}.

\textbf{Then, we compare all the submissions regarding unsupervised ASR on the hidden-set track}.
Unsupervised ASR is based on wav2vec 2.0-large, so it is reasonable to compare the two models. 
Unsupervised ASR improves PR, SID, and ER, only slightly degrading in ASR, SD, and ST.
 Unsupervised ASR has outstanding SID performance ranking at the 2nd in the Table of~\ref{table:exp}), only worse than WavLM-large.
 Moreover, in terms of the performance of ER, unsupervised ASR is only worse than WavLM-large and Unsupervised ASR plus T5.
 It is reasonable that by adding text information, unsupervised ASR can improve PR, but why it achieved good results on SID and ER is unclear and still under investigation. 
Adding T5 does not further improve any tasks except ER.

\section{Conclusions} 
\label{sec:conclusion}
\textit{SUPERB challenge at SLT 2022} challenges the speech community to build performant SSL techniques, and meanwhile achieve strong task generalizability and computation efficiency. The challenge aims to encourage researchers to consider more design criteria beyond model performance while building SSL techniques, such that SSL can expand its impact in more practical use cases. 
We review 14 models participating in this challenge, present our discoveries and highlight potential research directions in SSL.

\section{ACKNOWLEDGMENTS}
\label{sec:ack}
We thank National Center for High-performance Computing (NCHC) of National Applied Research
Laboratories (NARLabs) in Taiwan and Taiwan Web Service (TWS) for providing computational and storage resources used in this challenge.

\bibliographystyle{IEEEbib}
\bibliography{strings,refs}

\end{document}